\documentclass[letterpaper]{article} 
\usepackage{aaai25}  
\usepackage{times}  
\usepackage{helvet}  
\usepackage{courier}  
\usepackage[hyphens]{url}  
\usepackage{graphicx} 
\usepackage{natbib}  
\usepackage{caption} 
\usepackage{algorithm}
\usepackage{algorithmic}
\usepackage{color}
\usepackage{utfsym}
\usepackage{colortbl}
\usepackage{booktabs}
\usepackage{multirow}
\usepackage{multicol}
\usepackage{xspace}
\usepackage{amsfonts}
\usepackage{makecell}
\usepackage{array}
\usepackage{seqsplit}
\usepackage{amsmath}
\usepackage{mathrsfs}
\newcommand{\ourapproach}{ProtCLIP\xspace}
\newcommand{\ourdataset}{\textit{ProtAnno}\xspace}
\newcommand{\ourdatasetS}{\textit{ProtAnno-S}\xspace}
\newcommand{\ourdatasetD}{\textit{ProtAnno-D}\xspace}
\usepackage{newfloat}
\usepackage{listings}
\DeclareCaptionStyle{ruled}{labelfont=normalfont,labelsep=colon,strut=off} 
\floatstyle{ruled}
\newfloat{listing}{tb}{lst}{}
\floatname{listing}{Listing}
\nocopyright 

\title{\ourapproach: Function-Informed Protein Multi-Modal Learning}
\author{
    Hanjing Zhou\textsuperscript{\rm 1,2,3}\equalcontrib
    Mingze Yin\textsuperscript{\rm 1,2}\equalcontrib,
    Wei Wu\textsuperscript{\rm 4},
    Mingyang Li\textsuperscript{\rm 3},
    Kun Fu\textsuperscript{\rm 3}, \\
    Jintai Chen\textsuperscript{\rm 5}\thanks{Corresponding authors},
    Jian Wu\textsuperscript{\rm 2,6}\footnotemark[2],
    Zheng Wang\textsuperscript{\rm 3}\footnotemark[2]
}
\affiliations{
    \textsuperscript{\rm 1}College of Computer Science and Technology, Zhejiang University,  \\
    \textsuperscript{\rm 2}State Key Laboratory of Transvascular Implantation Devices of The Second Affiliated Hospital, School of Medicine, Zhejiang University, \\
    \textsuperscript{\rm 3}Alibaba Cloud Computing, \\
    \textsuperscript{\rm 4}School of Artificial Intelligence and Data Science, University of Science and Technology of China, \\
    \textsuperscript{\rm 5}AI Thrust, Information Hub, HKUST(Guangzhou), \\
    \textsuperscript{\rm 6}Zhejiang Key Laboratory of Medical Imaging Artificial Intelligence
    
    \{zhj85393, mzyin256, jtchen721\}@gmail.com, \\
    wujian2000@zju.edu.cn, \\ 
    \{sangheng.lmy, fukun.fu, wz388779\}@alibaba-inc.com, \\
    urara@mail.ustc.edu.cn
    
}

\usepackage{bibentry}

\begin{document}

\maketitle

\begin{abstract}
Multi-modality pre-training paradigm that aligns protein sequences and biological descriptions has learned general protein representations and achieved promising performance in various downstream applications. However, these works were still unable to replicate the extraordinary success of language-supervised visual foundation models due to the ineffective usage of aligned protein-text paired data and the lack of an effective function-informed pre-training paradigm. To address these issues, this paper curates a large-scale protein-text paired dataset called \ourdataset with a property-driven sampling strategy, and introduces a novel function-informed protein pre-training paradigm. Specifically, the sampling strategy determines selecting probability based on the sample confidence and property coverage, balancing the data quality and data quantity in face of large-scale noisy data. Furthermore, motivated by significance of the protein specific functional mechanism, the proposed paradigm explicitly model protein static and dynamic functional segments by two segment-wise pre-training objectives, injecting fine-grained information in a function-informed manner.
Leveraging all these innovations, we develop \ourapproach, a multi-modality foundation model that comprehensively represents function-aware protein embeddings.
On 22 different protein benchmarks within 5 types, including protein functionality classification, mutation effect prediction, cross-modal transformation, semantic similarity inference and protein-protein interaction prediction, our \ourapproach consistently achieves SOTA performance, with remarkable improvements of 75\% on average in five cross-modal transformation benchmarks, 59.9\% in GO-CC and 39.7\% in GO-BP 
protein function prediction. The experimental results verify the extraordinary potential of \ourapproach serving as the protein multi-modality foundation model.
\end{abstract}

\section{Introduction}
Proteins are essential functional units of cells, responsible for performing a wide range of vital and versatile functions crucial to life.
Mirroring the language-supervised pre-training paradigm towards powerful and unified vision representations~\cite{clip, DALLE2, ImageBind, BLIP2}, 
previous work has explored in the pre-training of multi-modality Protein Language Models (PLMs) by aligning protein sequences with textual function descriptions to achieve function-centric protein representations~\cite{OntoProtein, ProtST, ProteinCLIP, ProtET}.
However, these works were still unable to replicate the extraordinary success of image-text foundation models, and have shown to discard fine-grained protein functional information~\cite{ProteinCLIP}, which results in the suboptimal performance on cross-modal transformation~\cite{BioBridge} and localization prediction~\cite{ProtST}.
Literature has summarized that the success of visual foundation models primarily stems from \textbf{the efficient utilization of large-scale data}~\cite{clip, internVL} and \textbf{a holistic multi-modal pre-training framework}~\cite{KAD, PRIOR}, which points to two inherent obstacles that hinder further progress in multi-modal protein-biotext pre-training:

\noindent \textbf{(i) Absence of large-scale datasets and ineffective data usage.}
Large-scale aligned dataset is an indispensable part of obtaining powerful multi-modality foundation models. 
However, biotexts describing protein functions are much harder to construct than image captions, as often requiring detailed annotated process including manual review by experts or computational analysis by machines. 
This highlights the pressing need of large-scale multi-modal datasets containing protein sequences with high-quality functional annotations across multiple attribute domains.
Even with large-scale protein-biotext pairs, it is non-trivial to effectively inject biological property information into PLMs during multi-modal pre-training. This is primarily because the machine-analyzed process leads to numerous noisy labels (\textit{i.e.}, less accurate annotations)~\cite{Swiss_trEMBL}. Currently, there is still a lack of efficient learning techniques to effectively utilize large-scale proteins with noisy annotations for protein-biotext pre-training.


\noindent \textbf{(ii) Lack of a function-informed pre-training paradigm.}
Unlike the alignment of natural image-text pairs, the understanding of proteins is strongly influenced by their specific functional mechanism, which has been largely neglected by previous research yet. Proteins perform specific biological functions depending on their corresponding functional domains in 3D structural spaces. The amino acids at these active site are contiguous or discrete in 1D protein sequences. In this paper, we introduce the static and dynamic functional segment, new concepts which directly determine the specific protein functions and should be primarily focused during the alignment with biological function descriptions. However, we find existing protein-biotext pre-training works directly take after the original CLIP methodology for coarse-grained alignment, discarding the fine-grained information of protein unique functional mechanism (\textit{i.e.}, static or dynamic functional segments primarily determine protein specific functions and properties), which significantly prevents the better performance of protein-biotext pre-training.


Our work proposes a step towards constructing a universally applicable protein multi-modality foundation model aligning biological and natural language. We present \ourapproach, consistently alleviates the aforementioned two intrinsic problems and introduces remarkable innovations in multiple dimensions including the pre-training data, sampling strategy, and multi-modality objectives.

We first construct a high-quality protein-biotext paired dataset \ourdataset with sparse version (\ourdatasetS) and dense version (\ourdatasetD), derived from the existing protein function database~\cite{UniProt}. 
\ourapproach employs \ourdatasetD comprising 251.5 million aligned pairs for large-scale protein-biotext pre-training, which is the same order of magnitude as large-scale image-text pre-training. 
Since there exist some inevitable noisy annotations in \ourdatasetD (caused by machine-annotated bias), we propose a novel property-driven sampling strategy motivated by~\cite{MixMatch, DivideMix}.
Compared to the vanilla uniformly sampling, the proposed sampling strategy decides the selecting probability based on the sample confidence and property coverage, simultaneously balancing the data quality and data quantity in face of large-scale noisy labels.
Furthermore, a function-informed pre-training paradigm is constructed motivated by significance of the protein functional mechanism. 
Within such paradigm, we utilize CLIP loss~\cite{clip} to inject coarse-grained information, and two segment-wise objectives are designed to capture fine-grained information of the static and dynamic functional segments.
Concretely, on the one hand, we design a cross-modality reconstruction module to recover the masked static segments based on knowledge from both modalities.
On the other hand, the property prototype is exploited to aggregate dynamic segments in an unsupervised way. The resulting property-grouped dynamic segments are contrasted with property prototypes within the same protein-biotext pair, mitigating the mutual interference across multiple attribute domains.

Evaluated by extensive experiments, \ourapproach sets new state-of-the-art on 22 important yet challenging protein benchmarks within five types. For protein classification engineering and mutation effect prediction, the superiority of \ourapproach in representation learning attributes to incorporation of multi-modal information (\textit{e.g.}, 59.9\%/39.7\% improvements in Go-CC/GO-BP benchmarks). For cross-modal transformation, \ourapproach surpasses baselines by a significant margin (75\% improvement). 
For semantic similarity inference and protein-protein interaction prediction, \ourapproach ranks the best, which verifies effectiveness of the proposed data-efficient and function-informed multi-modal learning.



\section{Methods}
In this section, we first describe the curated multi-modal dataset, \ourdataset, and the property-driven sampling strategy to enhance data usage effectiveness. Next, we introduce the model architectures and our novel function-informed pre-training paradigm, which incorporates holistic multi-modal pre-training objectives to capture both coarse-grained and fine-grained information. Finally, we summarize the overall loss function used for protein-biotext pre-training.
\begin{table}[ht]
    \caption{Data distribution of \ourdatasetS and \ourdatasetD with different sample confidence. We highlight the confidence where protein entries are mostly concentrated in \textbf{bold}.}
    \vspace{-1em}
    \label{Tab:data_confidence}
    \begin{center}
    \resizebox{\linewidth}{!}{
    \begin{tabular}{c c c c c c}
            \toprule
            \textbf{Dataset} & \textbf{Conf-L1} & \textbf{Conf-L2} & \textbf{Conf-L3} & \textbf{Conf-L4} & \textbf{Conf-L5} \\
            \midrule
            \ourdatasetS & 0.1982 & 0.0980 & \textbf{0.6777} & 0.0229 & 0.0032 \\
            \midrule
            \ourdatasetD & 0.0013 & 0.0057 & 0.3269 & \textbf{0.6661} & 0.0000 \\
            \bottomrule
    \end{tabular}}
    \vspace{-1em}
    \end{center}
\end{table}

\begin{figure}[ht]
    \centering
    \includegraphics[width=0.95\linewidth]{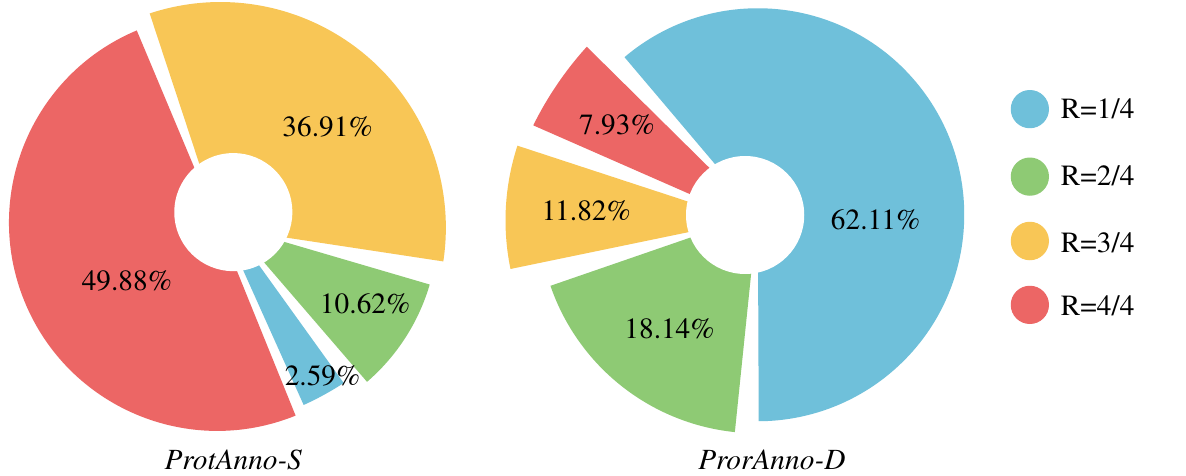}
    \caption{Data distribution of \ourdatasetS and \ourdatasetD with different property coverage.}
    \vspace{-1em}
    \label{Fig:data_coverage}
\end{figure}
\subsection{Pre-training data}
\label{Sec:data}
\begin{figure*}[htbp]
    \centering
    \includegraphics[width=0.97\textwidth]{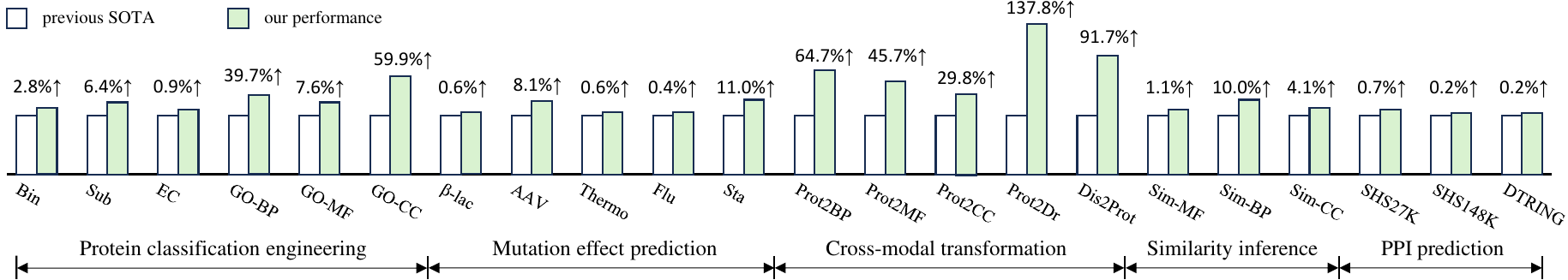}\vspace{-0.5em}
    \caption{Comparison results on 22 downstream benchmarks within five types. \ourapproach consistently achieves the state-of-the-art performance on all these tasks. PPI: protein-protein interaction.}
    \label{Fig:results}\vspace{-1em}
\end{figure*}
\subsubsection{Dataset Curation}
To enable pre-training of the protein multi-modality foundation model aligning biological and natural language, it is essential to build dataset containing large-scale pairs of protein sequences and textual property descriptions. Our pre-training data is sourced from SwissProt and trEMBL~\cite{Swiss_trEMBL}, containing proteins with textual descriptions. 
We align protein sequences with meticulously selected properties to curate \ourdataset, which is available in sparse version (\ourdatasetS) and dense version (\ourdatasetD). \ourdatasetS includes 0.5 million manually reviewed protein-biotext pairs with higher annotation quality, whereas \ourdatasetD comprises 251.5 million mostly computationally analyzed protein-biotext pairs which are less accurate due to the machine-annotated bias. 
To gain more insights into the dataset, we conduct extensive quantitative analyses, and display the compositional structure of \ourdataset with varying confidence $C$ and property coverage $R$ in Table~\ref{Tab:data_confidence} and Figure~\ref{Fig:data_coverage}. 
Details about selected properties for alignment and the data format are further illustrated in Appendix~\ref{App:data_format}.

\subsubsection{Property-driven Sampling Strategy}
For protein-biotext pre-training, most prior works only used scarce proteins with manually reviewed annotations (equivalent to \ourdatasetS), and the attempt to incorporate plentiful computationally analyzed proteins (equivalent to \ourdatasetD) has been unsuccessful, declaring ``\textit{data quality could be more important than data quantity.}''~\cite{ProtST}. However, we question and rethink this issue, and propose the property-driven sampling strategy which integrate the merits of the multi-modality data quality and data quantity. Specifically, the main considerations for sampling probability are sample confidence $C$ and property coverage $R$ and data size $N$. Note that the smaller the confidence, the more reliable the entry is, and $C \in \{1,2,3,4,5\}$, $R \in \{1/4, 2/4, 3/4, 4/4\}$.
Initially, we discard machine-annotated entries with $C=4,5$ (less accurate) and $R=1/4,2/4$ (low coverage) for comprehensive property understanding. 
Next, rather than uniform sampling, we explicitly build the sampling distribution according to the aforementioned three factors. 
The likelihood of selecting protein entries from cluster $u$ with $\{C_u, R_u, N_u\}$ during multi-modality pre-training is defined as:
\begin{equation}
    P = \frac{C_u^{-3} \cdot \sqrt{R_u} \cdot N_u}{\sum_{i,j,k} C_i^{-3} \cdot \sqrt{R_j} \cdot N_k}.
\end{equation}
In this paper, we perform large-scale protein-biotext pre-training exploiting \ourdatasetD, in conjunction with the proposed property-driven sampling strategy.

\subsection{Model Architecture}
\begin{figure*}[htbp]
    \centering
    \includegraphics[width=0.95\textwidth]{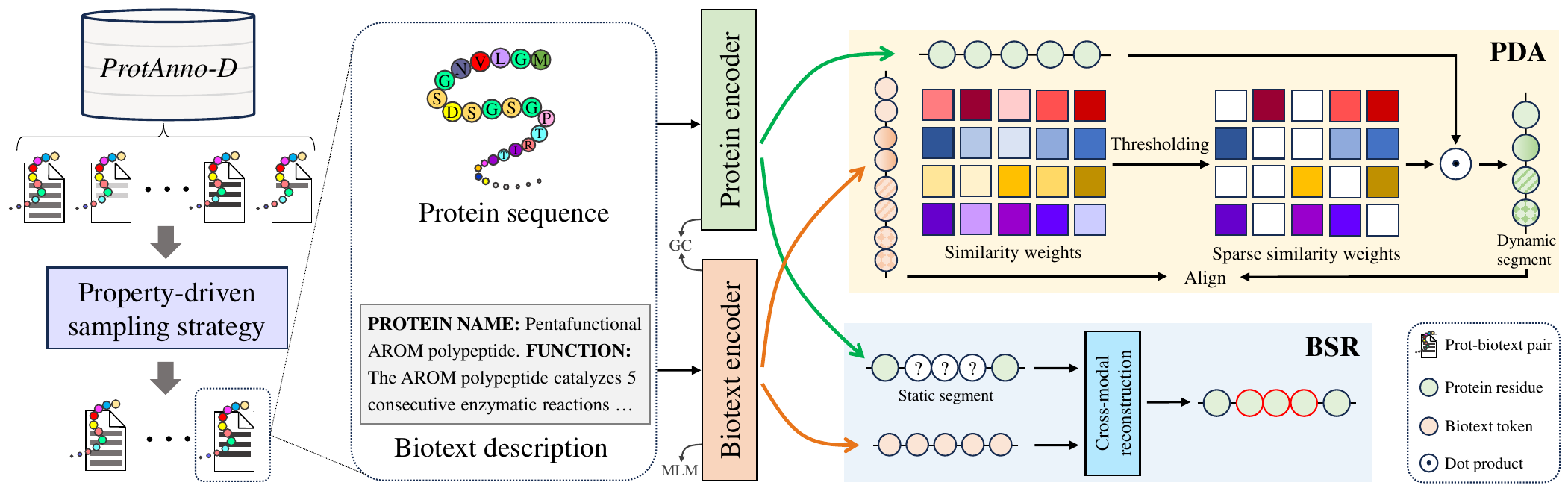}
    \caption{Overview of \ourapproach. We curate a large-scale protein-biotext dataset \ourdataset with a property-driven sampling strategy, and proposes a function-informed pre-training paradigm containing two segment-wise objectives BSR and PDA. }
    \vspace{-1em}
    \label{Fig:framework}
\end{figure*}

The overview of our framework is displayed in Figure~\ref{Fig:framework}, which contains a protein encoder and a biotext encoder. The protein encoder is a protein language model for learning biological features from protein sequences and we use pre-trained ESM-2-650M~\cite{esm-2} here. The biotext encoder is a text language model for learning linguistic features from biotext descriptions and we use PubMedBERT~\cite{PubMedBERT} here. Initialization with these two pre-trained large models significantly facilitates pre-training process by providing decent representations in the early stage of training.

\subsection{Function-informed Pre-training Paradigm}
To accomplish the holistic function-informed multi-modal pre-training, we jointly optimize four protein-biotext pre-training objectives, with two classic ones and two newly proposed segment-wise ones, customized for learning locality-aware and fine-grained information of protein specific functional mechanism. 

\subsubsection{Global Contrastive Loss}
Global Contrastive loss (GC) learning aligns representations of two modalities by encouraging positive pairs to have higher similarity in contrast to the negative pairs. Considering the effectiveness of $\textit{\textbf{L}}_{\text{GC}}$ for multi-modal understanding in many previous works~\cite{clip,BLIP2,MoSu} from different domains, we perform it to realize global alignment of protein-biotext. Given a batch of sequence-text pairs $\{(S_i, T_i)\}_{i=1}^{K}$, $\textit{\textbf{L}}_{\text{GC}}$ is composed of two symmetric standard InfoNCE loss:
\begin{equation}
\begin{split}
    \textit{\textbf{L}}_{\text{GC}} = -\frac{1}{2} \left[ \mathbb{E}_{p(S,T)}(log\frac{\exp(\mathtt{sim}(S_i, T_i)/\tau_1)}{\sum_{j=1}^{K} \exp(\mathtt{sim}(S_i, T_j)/\tau_1)}) \right. \\
    \left. +~\mathbb{E}_{p(S,T)}(log\frac{\exp(\mathtt{sim}(T_i, S_i)/\tau_1)}{\sum_{j=1}^{K} \exp(\mathtt{sim}(T_i, S_j)/\tau_1)}) \right ],
\end{split}
\end{equation}
where $\mathtt{sim}(;)$ is the consine similarity and $\tau_1$ denotes the temperature parameter that controls the softmax distribution.

\subsubsection{Biotext-guided Static Segment Reconstruction (BSR)}
Given the global contrastive objective modeling coarse-grained information, the fine-grained information of static and dynamic segments are ubiquitous, which primarily determines protein specific functions and properties. To capture such locality-aware information of static segments, we propose Biotext-guided Static segment Reconstruction (BSR) to reconstruct corrupted static segments using information from both modalities.
Specifically, given a sequence of protein residues $S=\{x_1, x_2,\dots,x_n\}$, we sample $l$ consecutive tokens as a static segment at a time, until the total sampling length reaches 15\% of $S$. In other words, we execute sampling iterations to prepare a random set of static segments $\{e_1, e_2, \dots, e_m\}$ with $e_i \in S$ for subsequent masking and reconstruction. 
At each iteration, we randomly select the starting point of each segment and its length $l$ follows a discrete uniform distribution between 5 and 10.
Note that all static segments are non-overlapping and their total length accounts for 15\% of $S$. 

Given the selected diverse static segments, we introduce a novel cross-modality reconstruction module to reconstruct masked segments according to the biotext functional descriptions, as displayed in Figure~\ref{Fig:framework}. Specifically, the protein sequence with masked segments $e^{m}$ and biotext $T$ are fed into a cross-attention module to obtain the fused representation by attending to all tokens along the biological property description. Then a MLP with the GELU activation~\cite{GELU} and layer normalization~\cite{LN} serves as the reconstruction head.
Formally, the loss function for BSR is:
\begin{equation}
    \textit{\textbf{L}}_{\text{BSR}} = \mathbb{E}_{p(T, e^{m})} H(\Phi(T, e^{m}), y_e),
\end{equation}
where $\Phi(T, e^{m})$ is the predicted probability of protein sequence with masked static segments $e^{m}$, and $y_e$ is the corresponding ground truth. $H(;)$ is the cross-entropy function. 

\subsubsection{Property-grouped Dynamic Segment Alignment (PDA)}
To capture the fine-grained information of dynamic segments, we propose Property-grouped Dynamic Segment Alignment (PDA), optimizing the alignment between property-grouped dynamic segments and corresponding property descriptions.

Specifically, a prototype memory bank is constructed to approximate property descriptive sentences, without any need to accurately retain redundant information such as syntax. 

Then the property prototype is exploited to aggregate dynamic segments in an unsupervised way, which are more flexible than static segments in BSR.  
Provided property description prototypes of biotext $T=\{a_1, a_2, a_3, a_4\}$ and the corresponding sequence of residues $S=\{x_1, x_2, \dots, x_n\}$, we first compute similarity weights as:
\begin{equation}
    w_{ij}=a_i \cdot x_j,~i=1,2,3,4,~j=1,2,\dots,n,
\end{equation}
where $w_{ij} \in \mathbb{R}$ and $\cdot$ is the inner product.
Then min-max normalization is applied along the residue dimension to normalize $w_{ij}$ to $[0,1]$. After that, some non-functional protein residues are discarded by sparsifying the similarity weights with a threshold $\theta$:
\begin{equation}
    \hat{w}_{ij}=
    \begin{cases}
        w_{ij}, & \text{if}~w_{ij} \geq \theta \\
        0, & \text{otherwise}.
    \end{cases}
\end{equation}
Eventually, we obtain the property-grouped dynamic segments by multiplying similarity weights and protein residues:
\begin{equation}
    e_i = \{\hat{w}_{ij} x_{j}~|~j=1,2,\dots,n\},~i=1,2,3,4.
\end{equation}
Property-grouped dynamic segment alignment is conducted to align these dynamic segments with property descriptions within the same protein-
biotext pair, mitigating the mutual interference across multiple attribute domains:

\begin{equation}
\begin{split}
    \textit{\textbf{L}}_{\text{PDA}} = -\frac{1}{2} \left[ \mathbb{E}_{p(e,a)}(log\frac{\exp(\mathtt{sim}(e_i, a_i)/\tau_2)}{\sum_{k} \exp(\mathtt{sim}(e_i, a_k)/\tau_2)}) \right. \\
    \left. +~\mathbb{E}_{p(e,a)}(log\frac{\exp(\mathtt{sim}(e_i, a_i)/\tau_2)}{\sum_{k} \exp(\mathtt{sim}(e_i, a_k)/\tau_2)}) \right ],
\end{split}
\end{equation}
where $\mathtt{sim}(;)$ represents the consine similarity and $\tau_2$ denotes the temperature parameter that controls the softmax distribution.

Aiming to extract the essential knowledge of protein sequences, we select the most relevant residues based on their similarities to each property description, resulting in segments of variable lengths. Owing to such variable length, dynamic segments are flexible to capture information of consecutive or non-consecutive functional residues, excluding redundant and non-functional ones. Additionally, the threshold $\theta$ directly influences the segment length by determining different number of zero values in each row of the similarity weights, which decouples similarities of individual residues to different property descriptions. In essence, the thresholding operation allows for different properties to match different residues that are the most relevant, thereby forming dynamic segments.

\subsection{Overall Loss Function}
The overall loss function of \ourapproach comprises four terms. Global contrastive loss $\textit{\textbf{L}}_{\text{GC}}$ learns coarse-grained information, while biotext-guided static segment reconstruction $\textit{\textbf{L}}_{\text{BSR}}$ and property-grouped dynamic segment alignment $\textit{\textbf{L}}_{\text{PDA}}$ focuses on fine-grained information. And we keep the protein masked language modeling $\textit{\textbf{L}}_{\text{MLM}}$ to preserve unimodal knowledge when injecting multi-modality information from biological texts.
We optimize these terms jointly via a weighted sum with hyper-parameters $\lambda_1$ and $\lambda_2$:
\begin{equation}
    \textit{\textbf{L}} = \textit{\textbf{L}}_{\text{GC}}+\lambda_1 \textit{\textbf{L}}_{\text{BSR}}+\lambda_2 \textit{\textbf{L}}_{\text{MLM}}+\textit{\textbf{L}}_{\text{PDA}}.
\end{equation}
During the training process, we observe a significant mutual interference between segment-level reconstruction $\textit{\textbf{L}}_{\text{BSR}}$ and token-level reconstruction $\textit{\textbf{L}}_{\text{MLM}}$, and set $\lambda_1+\lambda_2=1$. The investigation of their equilibrium is in Section~\ref{Sec:ablation}.

\section{Experiments}
\begin{figure*}[htbp]
    \centering
    \includegraphics[width=0.93\textwidth]{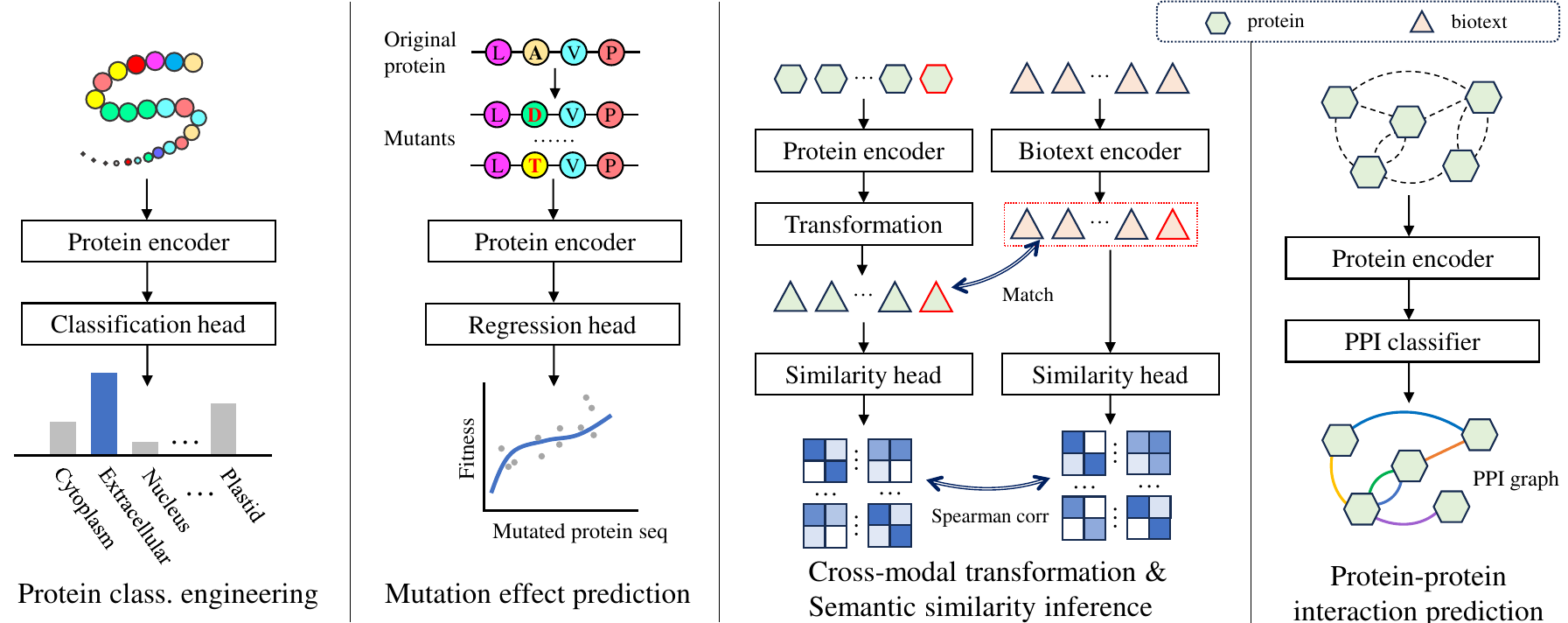}
    \vspace{-0.5em}
    \caption{An overview of downstream tasks within five types.}
    \label{Fig:downstream}
\end{figure*}

In this section, we first introduce some training setups, and then provide configurations and result discussions about five types of downstream applications (Figure~\ref{Fig:downstream}) on totally 22 benchmarks. Eventually, the analysis of ablation experiments are presented to further validate the effectiveness of our pre-training objectives.

\subsection{Training Setups}
We build our codes upon the PyTorch framework and conduct experiments on 64 Tesla V100 GPUs with 10,000 GPU hours.
An Adam optimizer is used (learning rate: $1.0 \times 10^{-5}$, weight decay: $0$) to train the model. The batch size is 2048 and 512 for pre-training and downstream experiments. 
Within the function-informed pre-training paradigm, we set hyper-parameters $\theta=0.3, \lambda_1=0.7, \lambda_2=0.3$.

\subsection{Protein Classification Engineering}
\label{Sec:exp1}
\subsubsection{Configurations}
Protein classification engineering aims to classify protein locations and functions. For location classification, we consider two such problems from DeepLoc~\cite{DeepLoc}, subcellular localization prediction (Sub) with 10 categories and binary localization prediction (Bin) with 2 categories. For function classification, we employ two benchmarks~\cite{DeepFRI} namely Enzyme Commission (EC) number prediction and Gene Ontology (GO) term prediction. On GO benchmark, there are three branches that predict molecular function (GO-MF), biological process (GO-BP) and cellular component (GO-CC).
The compared baselines include three parts: (a) four traditional protein encoders CNN~\cite{CNN}, ResNet~\cite{TAPE}, LSTM~\cite{TAPE}, Transformer~\cite{TAPE}; (b) 
four single-modal PLMs ProtBERT~\cite{ProtTrans}, OntoProtein~\cite{OntoProtein}, ESM-1b~\cite{esm-1b}, ESM2~\cite{esm-2}); (c) one multi-modal PLM ProtST-ESM2~\cite{ProtST}.
The evaluation metrics are accuracy for location prediction, and AUPR and F$_\text{max}$ for function prediction. AUPR denotes the pair-centric area under precision-recall curve. It computes average precision scores for all protein-biotext pairs, which is exactly the micro-average precision score for the classification problem. F$_\text{max}$ demotes the protein-centric maximum F-score.
\begin{table}[!ht]
    \caption{Results on location classification (Loc class) and mutation effect prediction (Effect pred) tasks. We highlight the best results in \textbf{bold}.}
    \vspace{-1em}
    \label{Tab:class1_reg}
    \begin{center}
    \setlength{\tabcolsep}{0.3em}
    \resizebox{\linewidth}{!}{
    \begin{tabular}{c|l|c c| c c c c c}
            \toprule
            \multirow{2}{*}{\textbf{\makecell{Modality \\ Amount}}}& \multicolumn{1}{c|}{\multirow{2}{*}{\textbf{Method}}} & \multicolumn{2}{c}{\textbf{Loc class} (\small{Acc \%})} & \multicolumn{5}{|c}{\textbf{Effect pred} (\small{Spearman's $\rho$})} \\
            \cmidrule{3-9}
            & & Bin & Sub & $\beta$-lac & AAV & Thermo & Flu & Sta \\
            \midrule
            \multicolumn{9}{c}{Tranditional models trained from scratch}\\
            \midrule
            \multirow{4}{*}{Single} & CNN & 82.67 & 58.73 & 0.781 & 0.746 & 0.494 & 0.682 & 0.637 \\
            & ResNet & 78.99 & 52.30 & 0.152 & 0.739 & 0.528 & 0.636 & 0.126 \\
            & LSTM & 88.11 & 62.98 & 0.139 & 0.125 & 0.564 & 0.494 & 0.533 \\
            & Transformer & 75.74 & 56.02 & 0.261 & 0.681 & 0.545 & 0.643 & 0.649 \\
            \midrule
            \multicolumn{9}{c}{PLMs under linear probing} \\
            \midrule
            \multirow{4}{*} {Single} & ProtBERT & 81.54 & 59.44 & 0.616 & 0.209 & 0.562 & 0.339 & 0.697 \\
            & OntoProtein & 84.87 & 68.34 & 0.471 & 0.217 & 0.605 & 0.432 & 0.688 \\
            & ESM-1b & 91.61 & 79.82 & 0.528 & 0.454 & 0.674 & 0.430 & 0.750 \\
            & ESM2 & 91.32 & 80.84 & 0.559 & 0.374 & 0.677 & 0.456 & 0.746 \\
            \midrule
            \multirow{2}{*}{Multiple} & ProtST-ESM2 & 92.52 & 83.39 & 0.565 & 0.398 & 0.681 & 0.499 & 0.776 \\
            & \textbf{\ourapproach} & \cellcolor{gray!20}94.39 & \cellcolor{gray!20}83.65 & \cellcolor{gray!20}0.565 & \cellcolor{gray!20}0.532 & 
            \cellcolor{gray!20}0.682 & 
            \cellcolor{gray!20}0.503 & 
            \cellcolor{gray!20}0.795 \\
            \midrule
            \multicolumn{9}{c}{PLMs under full tuning}\\
            \midrule
            \multirow{4}{*}{Single}
            & ProtBERT & 91.32 & 76.53 & 0.731 & 0.794 & 0.660 & 0.679 & 0.771 \\
            & OntoProtein & 92.47 & 77.59 & 0.757 & 0.791 & 0.662 & 0.630 & 0.731 \\
            & ESM-1b & 92.40 & 78.13 & 0.839 & 0.821 & 0.669 & 0.679 & 0.694 \\
            & ESM2 & 91.72 & 78.67 & 0.867 & 0.817 & 0.672 & 0.677 & 0.718 \\
            \midrule
            \multirow{2}{*}{Multiple}& ProtST-ESM2 & 92.52 & 80.22 & 0.879 & 0.825 & 0.682 & 0.682 & 0.738 \\
            & \textbf{\ourapproach} & \cellcolor{gray!20}\textbf{95.08} & \cellcolor{gray!20}\textbf{85.34} & \cellcolor{gray!20}\textbf{0.884} & \cellcolor{gray!20}\textbf{0.892} & 
            \cellcolor{gray!20}\textbf{0.686} & 
            \cellcolor{gray!20}\textbf{0.685} & 
            \cellcolor{gray!20}\textbf{0.819} \\
            \bottomrule
    \end{tabular}}
    \end{center}
\end{table}
\begin{table}[!ht]
    \caption{Results on function classification task. We highlight the best results in \textbf{bold}.}
    \vspace{-1em}
    \label{Tab:class2}
    \begin{center}
    \setlength{\tabcolsep}{0.3em}
    \resizebox{\linewidth}{!}{
    \begin{tabular}{c|l|c c| c c| c c| c c}
            \toprule
            \multirow{2}{*}{\textbf{\makecell{Modality \\ Amount}}} & \multicolumn{1}{c|}{\multirow{2}{*}{\textbf{Method}}} & \multicolumn{2}{c|}{\textbf{EC}} & \multicolumn{2}{c|}{\textbf{GO-BP}} & \multicolumn{2}{c|}{\textbf{GO-MF}} & \multicolumn{2}{c}{\textbf{GO-CC}}\\
            \cmidrule{3-10}
            & & \small{AUPR} & $F_{\text{max}}$ & \small{AUPR} & $F_{\text{max}}$ & \small{AUPR} & $F_{\text{max}}$ & \small{AUPR} & $F_{\text{max}}$ \\
            \midrule
            \multicolumn{10}{c}{Traditional model trained from scratch}\\
            \midrule
            \multirow{4}{*}{Single} & CNN & 0.540 & 0.545 & 0.165 & 0.244 & 0.380 & 0.354 & 0.261 & 0.387 \\
            & ResNet & 0.137 & 0.187 & 0.166 & 0.280 & 0.281 & 0.267 & 0.266 & 0.403 \\
            & LSTM & 0.032 & 0.082 & 0.130 & 0.248 & 0.100 & 0.166 & 0.150 & 0.320 \\
            & Transformer & 0.187 & 0.219 & 0.135 & 0.257 & 0.172 & 0.240 & 0.170 & 0.380 \\
            \midrule
            \multicolumn{10}{c}{PLMs under full tuning}\\
            \midrule
            \multirow{4}{*}{Single}
            & ProtBERT & 0.859 & 0.838 & 0.188 & 0.279 & 0.464 & 0.456 & 0.234 & 0.408 \\
            & OntoProtein & 0.854 & 0.841 & 0.284 & 0.436 & 0.603 & 0.631 & 0.300 & 0.441 \\
            & ESM-1b & 0.884 & 0.869 & 0.332 & 0.452 & 0.630 & 0.659 & 0.324 & 0.477 \\
            & ESM2 & 0.888 & 0.874 & 0.340 & 0.472 & 0.643 & 0.662 & 0.350 & 0.472 \\
            \midrule
            \multirow{2}{*}{Multiple}& ProtST-ESM2 & 0.898 & 0.878 & 0.342 & 0.482 & 0.647 & 0.668 & 0.364 & 0.487 \\
            & \textbf{\ourapproach} & \cellcolor{gray!20}\textbf{0.906} & \cellcolor{gray!20}\textbf{0.908} & \cellcolor{gray!20}\textbf{0.567} & \cellcolor{gray!20}\textbf{0.574} & 
            \cellcolor{gray!20}\textbf{0.696} & 
            \cellcolor{gray!20}\textbf{0.691} & 
            \cellcolor{gray!20}\textbf{0.582} &
            \cellcolor{gray!20}\textbf{0.541} \\
            \bottomrule
    \end{tabular}}\vspace{-1em}
    \end{center}
\end{table}
\subsubsection{Results}
Table~\ref{Tab:class1_reg} (left) and Table~\ref{Tab:class2} show that \ourapproach establishes state-of-the-art results on all six classification benchmarks under both linear probing and full tuning settings.
Moreover, \ourapproach performs best on protein classification engineering among all five type of downstream tasks.
\begin{table*}[!ht]
\begin{minipage}{0.49\textwidth}
    \caption{Mean reciprocal rank (MRR) results on cross-modal transformation task. Prot: protein.}\vspace{-1em}
    \label{Tab:trans}
    \setlength{\tabcolsep}{0.3em}
    \begin{center}
    \resizebox{1\linewidth}{!}{
    \begin{tabular}{l c c c c c}
            \toprule
            \multicolumn{1}{c}{\textbf{Method}} & \textbf{Prot2BP} & \textbf{Prot2MF} & \textbf{Prot2CC} & \textbf{Prot2Drug} & \textbf{Disease2Prot} \\
            \midrule
            ComplEx & 0.084 & 0.100 & 0.099 & 0.079 & 0.059 \\
            DistMult & 0.054 & 0.089 & 0.095 & 0.044 & 0.033 \\
            RotatE & 0.079 & 0.119 & 0.107 & 0.125 & 0.070 \\
            BioBridge & 0.136 & 0.326 & 0.319 & 0.172 & 0.084 \\
            \ourapproach & \cellcolor{gray!20}\textbf{0.224} & \cellcolor{gray!20}\textbf{0.475} & \cellcolor{gray!20}\textbf{0.414} & \cellcolor{gray!20}\textbf{0.409} & \cellcolor{gray!20}\textbf{0.161} \\
            \bottomrule
    \end{tabular}}\vspace{-1em}
    \end{center}
\end{minipage}
\hfill
\begin{minipage}{0.49\textwidth}
    \caption{Results on semantic similarity inference (Sim) and protein-protein interaction prediction (PPI) tasks.}\vspace{-1em}
    \label{Tab:sim_PPI}
    \begin{center}
    \resizebox{0.96\linewidth}{!}{
    \begin{tabular}{l| c c c| c c c c}
            \toprule
            \multicolumn{1}{c|}{\multirow{2}{*}{\textbf{Method}}} & \multicolumn{3}{c|}{\textbf{Sim} (\small{Spearman's $\rho$})} & \multicolumn{3}{c}{\textbf{PPI} (\small{F1 score})}  \\
            \cmidrule{2-7}
            & MF & BP & CC & SHS27K & SHS148K & STRING \\
            \midrule
            ESM2-3B & 0.33 & 0.42 & 0.23 & 0.732 & 0.733 & 0.834 \\
            KeAP & 0.41 & 0.41 & 0.40 & 0.733 & 0.726 & 0.834 \\
            BioBridge & 0.91 & 0.80 & 0.73 & 0.739 & 0.739 & 0.836 \\
            \ourapproach & \cellcolor{gray!20}\textbf{0.92} & \cellcolor{gray!20}\textbf{0.88} & \cellcolor{gray!20}\textbf{0.76} & \cellcolor{gray!20}\textbf{0.744} & \cellcolor{gray!20}\textbf{0.740} & \cellcolor{gray!20}\textbf{0.838} \\
            \bottomrule
    \end{tabular}}\vspace{-1em}
    \end{center}
\end{minipage}
\end{table*}
\subsection{Mutation Effect Prediction}
\subsubsection{Configurations}
Mutation effect prediction is a regression task that predicts the effect of residue mutations on protein fitness. We utilize $\beta$-lactamase ($\beta$-lac) landscape from PEER~\cite{PEER}, Fluorescence (Flu) and Stability (Sta) landscapes from TAPE~\cite{TAPE}, and AAV and Thermostability (Thermo) landscapes from FLIP~\cite{FLIP}. Following the setup of~\cite{ProtST}, we use ``two vs many'' and ``human cell'' dataset splits for AAV and Thermo, and the split settings of the rest tasks remain default and unchanged.
The baselines remain the same as mentioned in Section~\ref{Sec:exp1}. The performance is measured by Spearman’s $\rho$. 
Moreover, 
we evaluate \ourapproach and PLMs under both linear probing and full tuning settings on location prediction and mutation effect prediction tasks.

\subsubsection{Results}
Table~\ref{Tab:class1_reg} illustrates that \ourapproach consistently ranks the best among other baselines. We can observe that although traditional models (\textit{e.g.}, CNN) pose strong competition in mutation effect prediction, \ourapproach still retains the lead, especially on Stability benchmark in full tuning setting.

\subsection{Cross-modal Transformation}
\subsubsection{Configurations}
Cross-modal transformation matches the transformed embedding with candidates from the target modality, where embeddings from \ourapproach are transformed by an extra transformation module. Following~\cite{BioBridge}, we leverage the raw knowledge graph (KG) data and undertake some preprocessing steps, with the training/validation/test split of 80\%/10\%/10\%. The baselines are BioBridge~\cite{BioBridge} and three knowledge graph embedding methods 
(ComplEx~\cite{ComplEx}, DistMult~\cite{DistMult}, RotatE~\cite{RotatE}).
We use mean reciprocal rank (MRR) as the metric.

\subsubsection{Results}
Table~\ref{Tab:trans} reports our remarkable enhancement over all baselines. The first three baselines are traditional KG encoders trained from scratch, which lack flexibility, while BioBridge cannot fully unleash the potential of PLMs. Instead, \ourapproach compensates for their shortcomings and incorporates flexibility, data-efficiency and high performance.
Particularly, \ourapproach is 2.4 $\times$ better than the best baseline for ``Prot2Drug'' and 2 $\times$ better for ``Prot2BP'' and ``Disease2Prot'', which signals the superiority of \ourapproach in multimodal understanding.

\subsection{Semantic Similarity Inference}
\label{Sec:exp4}
\subsubsection{Configurations}
Semantic similarity inference computes the relevance between predicted and groundtruth similarity matrices~\cite{exp_sim}. our goal is to evaluate the extent to which the encoded protein embeddings can capture biomolecular functional similarity (\textit{i.e.}, BP, CC, MF). The predicted matrix contains pairwise Manhattan Similarities of the encoded protein embeddings, while the groundtruth stores pairwise Lin Similarities of the protein associated BP, MF, and CC. We compare \ourapproach with three baselines (\textit{i.e.}, ESM2-3B~\cite{esm-2}, KeAP~\cite{KeAP}, BioBridge~\cite{BioBridge}). The metric is Spearman's $\rho$.
\subsubsection{Results}
In Table~\ref{Tab:sim_PPI} (left), \ourapproach achieves the best performance over other baselines. In particular, \ourapproach 
surpasses the vanilla ESM2-3B by a large margin, demonstrating the proposed data-efficient and function-informed multi-modal learning is generally beneficial to the unimodal PLM.

\subsection{Protein-Protein Interaction Prediction}
\subsubsection{Configurations}
Protein-protein interaction (PPI) prediction seeks to classify 7 interaction types of a pair of proteins. Following~\cite{OntoProtein}, we extract the protein embeddings with \ourapproach and baselines, which serve as the input for a graph neural network model to be trained on the PPI network. The
baselines remain the same as mentioned in Section~\ref{Sec:exp4}. Additionally, F1 score is reported on SHS27K~\cite{SHSdataset}, SHS148K~\cite{SHSdataset} and STRING~\cite{STRING} datasets for evaluation.
\subsubsection{Results}
Table~\ref{Tab:sim_PPI} (right) presents average results on three benchmarks. \ourapproach performs the best and exceeds the prior state-of-the-art BioBridge owing to its pre-training on the enormous dataset \ourdatasetD with the property-driven sampling strategy.

\subsection{Ablation Study}
\label{Sec:ablation}
We conduct extensive ablation experiments from multiple aspects. Unless otherwise specified, ESM-2-150M serves as the protein encoder and we evaluate on three downstream benchmarks from different types in ablation experiments.

\subsubsection{Ablation study on Pre-training Data}
As seen in Section~\ref{Sec:data}, we curate a new dataset \ourdataset with a property-driven sampling strategy. Table~\ref{Tab:ablation_data} displays comparison of different pre-training data organization. 
Obviously, single dataset pre-training and pretrain+finetune (first pretrained on machine-annotated data, then fine-tuned on manually-reviewed data) are inferior to the model pre-trained on \ourdatasetD with the proposed sampling strategy. Such phenomenon demonstrates that low-quality data still holds potential value if subjected to elaborate processing and sampling, and \ourdataset strikes a good balance between data quality and data quantity.

\subsubsection{Ablation Study on Pre-training Objectives}
Table~\ref{Tab:ablation_loss} reports results with full or partial pre-training objectives. We can observe that  both PDA and BSR are essential for injecting fine-grained information, and the absence of PDA leads to a more significant drop compared to the lack of BSR. Such results signal the competence of our function-informed paradigm for protein-biotext multi-modal learning.
\begin{table}[!ht]
    \caption{Analysis on pre-training data. Pretrain+fintune: first pretrained on low accurate data, then fine-tuned on high accurate data. Property-driven sampling strategy: pretrained on \ourdatasetD with the proposed sampling strategy.}
    \vspace{-1em}
    \label{Tab:ablation_data}
    \begin{center}
    \resizebox{0.9\linewidth}{!}{
    \begin{tabular}{l|c | c  c | c}
            \toprule
            \multicolumn{1}{c|}{\multirow{2}{*}{\textbf{\makecell{Pre-training \\ data}}}} & \textbf{Sub} & \multicolumn{2}{c|}{\textbf{EC}} & \textbf{Prot2MF} \\
            \cmidrule{2-5}
            & Acc \% & \small{AUPR} & $F_{\text{max}}$ & \small{MRR} \\
            \midrule
            \ourdatasetS & 72.41 & 0.216 & 0.282 & 0.246 \\
            \ourdatasetD & 73.72 & 0.282 & 0.309 & 0.256 \\
            Pretrain+finetune & 74.98 & 0.312 & 0.404 & 0.283 \\
            \midrule
            Our sampling strategy & \textbf{75.77} & \textbf{0.384} & \textbf{0.441} & \textbf{0.299} \\
            \bottomrule
    \end{tabular}}
    \vspace{-1em}
    \end{center}
\end{table}
\begin{table}[!ht]
    \caption{Ablation study on pre-training objectives.}
    \vspace{-1em}
    \label{Tab:ablation_loss}
    \begin{center}
    \resizebox{0.85\linewidth}{!}{
    \begin{tabular}{l|c | c c | c}
            \toprule
            \multicolumn{1}{c|}{\multirow{2}{*}{\textbf{Config}}} & \textbf{Sub} & \multicolumn{2}{c|}{\textbf{EC}} & \textbf{Prot2MF} \\
            \cmidrule{2-5}
            & Acc \% & \small{AUPR} & $F_{\text{max}}$ & \small{MRR} \\
            \midrule
            w/o $\textit{\textbf{L}}_{\text{BSR}}$ & 76.09 & 0.189 & 0.254 & 0.282 \\
            w/o $\textit{\textbf{L}}_{\text{PDA}}$ & 73.64 & 0.136 & 0.227 & 0.210 \\
            \midrule
            Full loss & \textbf{76.52} & \textbf{0.204} & \textbf{0.320} & \textbf{0.312} \\
            \bottomrule
    \end{tabular}}\vspace{-1em}
    \end{center}
\end{table}
\begin{figure}[t]
    \centering
    \includegraphics[width=0.9\linewidth]{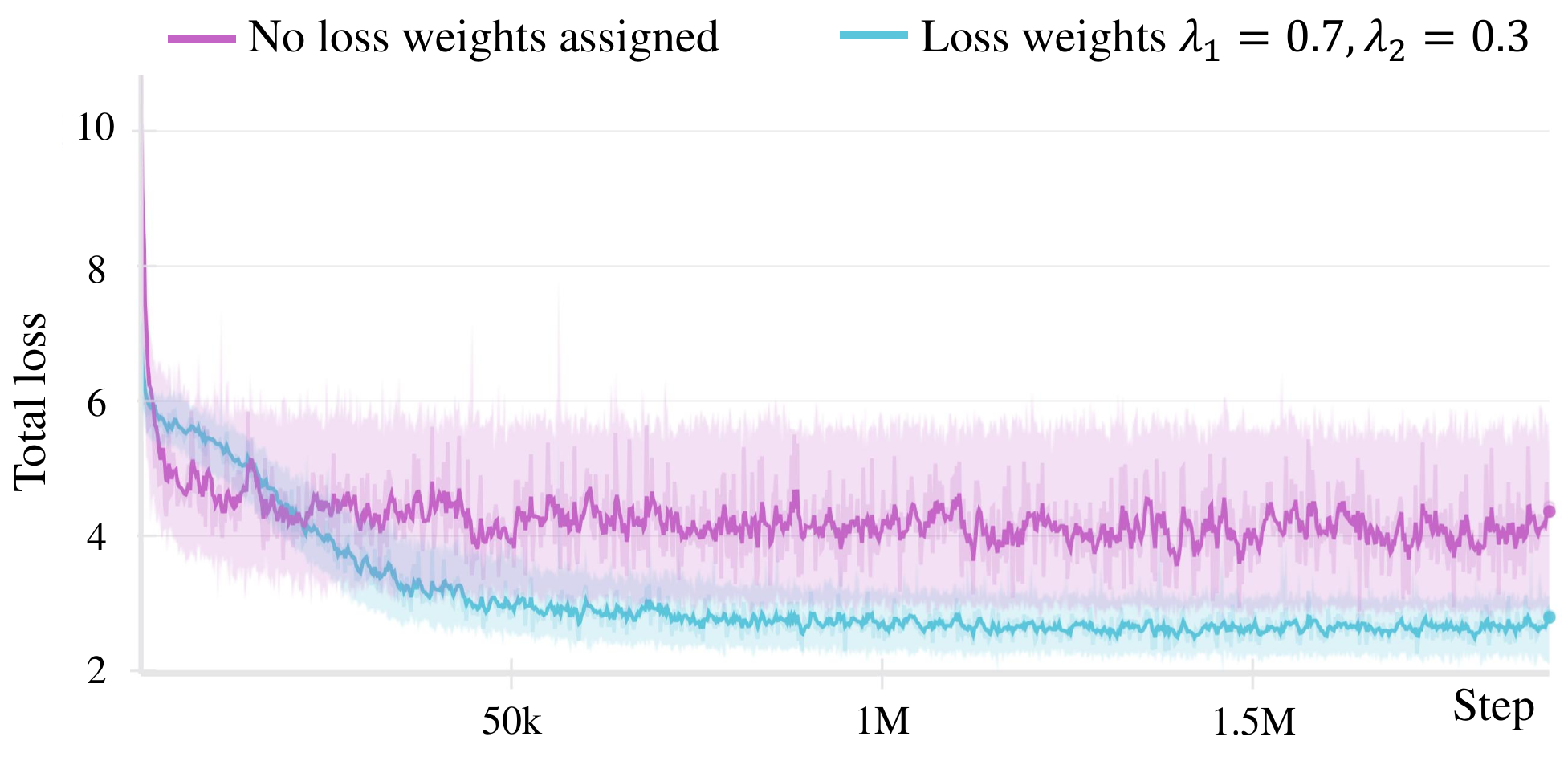}\vspace{-1em}
    \caption{Mutual interference between segment-wise reconstruction $\textbf{\textit{L}}_\text{BSR}$ and token-wise reconstruction $\textbf{\textit{L}}_\text{MLM}$ during the pre-training process.}
    \vspace{-1em}
    \label{Fig:interference}
\end{figure}
\begin{figure}[t]
    \centering
    \includegraphics[width=0.97\linewidth]{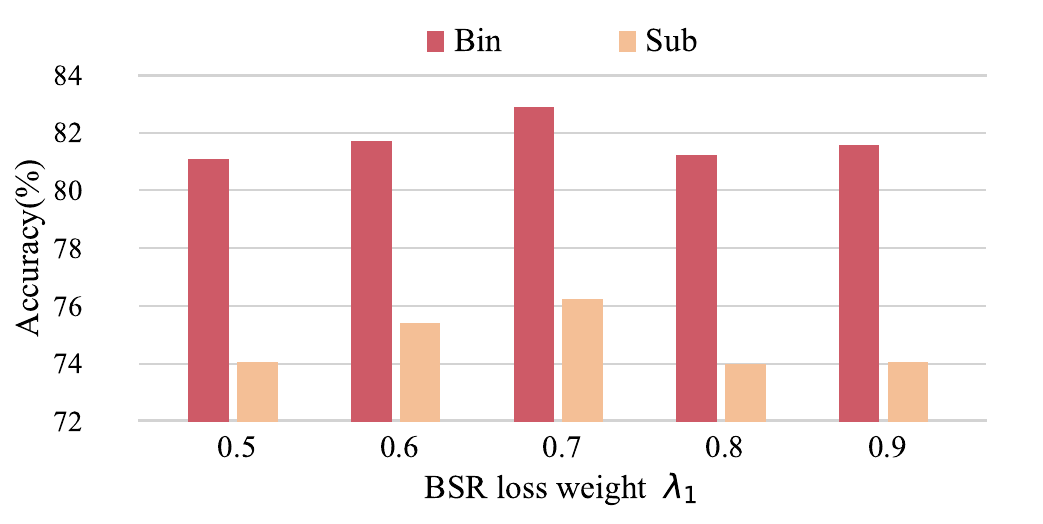}\vspace{-1em}
    \caption{Ablation study on loss weights.}
    \vspace{-1em}
    \label{Fig:ablation_weights}
\end{figure}
\subsubsection{Ablation Study on Loss Weights}
During the pre-training process, we observe a significant mutual
interference between segment-level reconstruction $\textbf{\textit{L}}_\text{BSR}$ and token-level reconstruction $\textbf{\textit{L}}_\text{MLM}$. As depicted in Figure~\ref{Fig:interference}, the loss curve fluctuate violently without falling if we apply no loss weights whereas it declines smoothly if the appropriate loss weights are assigned. Therefore, it is essential to find the optimal loss wights.
In Figure~\ref{Fig:ablation_weights}, different values of loss weights $\lambda_1$ yield different ablation results on two location classification benchmarks. Due to evident advantages, the ultimate weights are $\lambda_1=0.7$ and thus $\lambda_2=1-\lambda_1=0.3$.

\section{Related Work}
\subsection{Multi-modal Image-Text Pre-training}
In an effort to overcome the limitations of single-modality learning~\cite{groupon}, multi-modal image-text pre-training has been introduced to learn and align visual and textual representations by pre-training the model on large-scale image-text pairs. 
One of the most representative methods is CLIP~\cite{clip} , which has achieved multi-modal alignment through contrastive learning with massive noisy data scrapped from the internet. BLIP-family~\cite{BLIP, BLIP2} has continuously explored image-text unification and model lightweighting. Aiming to reduce the modality gap, LaVIT~\cite{LaVIT} and FDT~\cite{FDT} have respectively designed a unified discrete tokenizer to embed visual and textual input.
Large-scale image-text pre-training has become a widely used paradigm for learning general vision representations for a wide range of downstream tasks as well as for constructing multi-modality foundation models~\cite{LLaVA, LLaVA-Med}. 
Despite their impressive performance, previous methods have only learned coarse-grained representations and ignored localized details.
Motivated by this, SPARC~\cite{SPARC} proposes a fine-grained sequence-wise loss., encoding detailed information in a computationally inexpensive way. 
FILIP~\cite{filip} has constructed a cross-modal late interaction mechanism to optimize the token-wise maximal similarity between image and text. 
MGCA~\cite{MGCA}, PRIOR~\cite{PRIOR} and MedKLIP~\cite{MedKLIP} learn complex semantic structures in medical reports from different levels (disease-level, sentence-level and entity-level).
However, most of them are specifically tailored for image-text alignment, and cannot seamlessly be applied to multi-modal protein-biotext pre-training.

\subsection{Multi-modal Protein-Biotext Pre-training}
Recently, models that jointly pre-train protein sequences and biotext descriptions have gradually drawing the attention of researchers. OntoProtein~\cite{OntoProtein} first incorporates knowledge graphs to enhance protein representation with external biological descriptions. 
Chroma~\cite{chroma} conducts text-guided protein backbone editing towards desired properties and functions. Meanwhile, ProtDT~\cite{ProtDT} is a newly proposed multi-modal framework that aligns the representations of proteins and biotexts, and leverages textual descriptions 
for protein design. ProtST~\cite{ProtST} has shown a tremendous performance on exploiting biomedical function annotations to enhance protein sequence understanding. Additionally, a novel multi-modal framework for the accurate prediction of protein functional descriptions in free text format is proposed by~\cite{Prot2Text}. BioBridge~\cite{BioBridge} introduces a bridge module to learn transformations between protein, molecule and biotext foundation models. Nevertheless, existing works of protein-biotext alignment primarily exploit the global alignment objective proposed by CLIP~\cite{clip}, without utilizing protein specific functional mechanism to fully facilitate fine-grained understanding of protein and biotext.

\section{Conclusion}
This paper has accomplished data-efficient and function-informed multi-modal learning of proteins and biotexts. We build the \ourdataset dataset with large-scale aligned protein sequences and functional descriptions. The property-driven sampling strategy is introduced to strike a balance between data quality and data quantity for pre-training, thereby facilitating the effective harnessing of large-scale noisy data. Inspired by the intricate mechanisms of protein functionality, we novelly adopt a function-informed pre-training paradigm with newly proposed segment-wise objectives to explicitly model protein static and dynamic segments. Such paradigm seamlessly integrates multi-modality information from coarse-grained to fine-grained levels, culminating in the holistic function-centric protein representation. We also identified that \ourapproach achieves the new state-of-the-art results on 22 protein downstream benchmarks. In the future, we envision that \ourapproach has the potential to serve as the protein multi-modality foundation model to promote controllable protein discovery and optimization in real-world scenarios. 

\section{Acknowledgments}
This research was partially supported by National Natural Science Foundation of China under grants No. 12326612, Zhejiang Key R\&D Program of China under grant No. 2023C03053, the Opening Foundation of the State Key Laboratory of Transvascular Implantation Devices, grant No. SKLTID2024003, and Alibaba Research Intern Program.

\bibliography{references}

\clearpage
\appendix
\section{Protein-Biotext Data Format}
\label{App:data_format}
Here we present the data format of our multi-modal pre-training dataset, containing aligned pairs of the protein sequence and the biotext description. The protein sequence is a set of amino acids with the specific linear arrangement. The biotext description is annotations of multiple protein properties, from which we select four objects for pre-training considering their real-world value, coverage, etc. Informed by~\cite{ProtST}, the introduction of four objects is as follows: (1) ``Protein Name'': the full protein name recommended by the UniProt consortium~\cite{UniProt}; (2) ``Function'': diverse functions owned by a protein; (3) ``Subcellular Location'': the location and topology of a mature protein in the cell; (4) ``Similarity'': information about the protein families that a protein belongs to. These objects above are concatenated with a space delimiter to form the text input for pre-training, omitting any that are missing (mentioned in Section). And we add annotation prefixes in front of property descriptions: ``PROTEIN NAME'' for the protein name property; ``FUNCTION'' for the protein function property; ``SUBCELLULAR LOCATION'' for the subcellular location property; ``SIMILARITY'' for the protein family property. As illustrated in Table~\ref{Tab:data_example}, we present several property descriptions coupled with the corresponding entry names and protein sequences from \ourdataset. Note that there are some cases where four property descriptions are incomplete.

\begin{table*}[!ht]
    \caption{Examples of our pre-training data format. Note that pre-training input is a sequence-biotext pair, and we add the extra ``Entry name'' in this table for better indexing and understanding.}
    
    \label{Tab:data_example}
    \begin{center}
    \begin{tabular}{p{9em}|p{25em}|p{16em}}
            \toprule
            \textbf{Entry name} & \textbf{Protein sequence} & \multicolumn{1}{c}{\textbf{Biotext description}} \\
            \midrule
            A0A010SAB3$\underline{\hspace{0.5em}}$9PEZI & \seqsplit{MANSPHGGVLKDLFARDAPRQSELFAEADKLPSLLLTERHLCDLELILNGGFSPLEGFMTEKDYNGVVKDNRLADGNLFSMPITLDVSQQQIDTLSIKPGARITLRDLRDDRNLAILTVEDVYKPDRVKEAIEVFGSDDDTHPGVKHLFNNTNDFYVGGKLEAIQRLAHYDFLDLRFTPAELRQHFEKLGWNKVVAFQTRNPMHRAHRELTVRAARSQQANVLIHPVVGMTKPGDIDHFTRVRVYKALLPRYPNGMAALALLPLAMRMGGPREAIWHAIIRKNHGATHFIVGRDHAGPGKNKNGKDHYGPYDAQVAVQKYSDELGITMVEFQEMIYIPDRDEYQPANEIAPGTHTANISGTELRNRLKTGKEIPAWFSYPEVVKVLREQNPLPAQKGFTIFLTGLLNSGKDQIAKALQVTLNQGGGRSVSLLLGETVRHELSSELGFSREDRDKNVGRIAFVASELTRAGAAVIAAPIAPFDEARQKARELVEKAGPFFLVHVATPLEYAEKTDKRGIYQKARNGDIKGFTGVDDPYEAPAKADLVVNLEQQTVRSIVHQIVLLLESQGLLDRL} & \textbf{PROTEIN NAME:} Sulfate adenylyltransferase. \textbf{FUNCTION:} Catalyzes the first intracellular reaction of sulfate assimilation, forming adenosine-5'-phosphosulfate (APS) from inorganic sulfate and ATP. Plays an important role in sulfate activation as a component of the biosynthesis pathway of sulfur-containing amino acids. \textbf{SUBCELLULAR LOCATION:} Cytoplasm. \textbf{SIMILARITY:} Belongs to the APS kinase family. \\
            \midrule
            A0A009GHC8$\underline{\hspace{0.5em}}$ACIBA & \seqsplit{MDIFPISLKLQQQRCLIVGGGHIALRKATLLAKAGAIIDVVAPAIEDQLLQLITTTGGVSFIEAFTEKFLSTPYRLVIAATNDAEVNKTVFEQCEARNLLVNSVDDIPHCRFMVPAIIDRSPLIVSVASNGTSPVLSRQIRTQLETSIPHGMGKLAEFSGKWRNQVKEKISNPDERRIFWENLYASPLKEQVFNDNLDVADSMLEQALQEWKAPKGEVYLVGAGPGDPELITLKALRLMQQADVVIYDRLVSAPILELCRRDATKIYVGKARSNHSVPQEGINALLVDYAKKGKRVCRLKGGDPFIFGRGGEEIQELFQAGVPFQVVPGITAASGCSAYAGIPLTHRDYAQSVRFLTGHLKEGSPELPWNELVYENQTLVLYMGLVGLERICEQLIAHGQRPDMPVALISKGTTPEQKVVVGSLADIASKVTEHQIHAPTLTIIGEVVRLREQLQWN} & \textbf{PROTEIN NAME:} Siroheme synthase. \textbf{FUNCTION:} Multifunctional enzyme that catalyzes the SAM-dependent methylations of uroporphyrinogen III at position C-2 and C-7 to form precorrin-2 via precorrin-1. Then it catalyzes the NAD-dependent ring dehydrogenation of precorrin-2 to yield sirohydrochlorin. Finally, it catalyzes the ferrochelation of sirohydrochlorin to yield siroheme. \textbf{SIMILARITY:} Belongs to the precorrin methyltransferase family. \\
            \midrule
            A0A024R324$\underline{\hspace{0.5em}}$HUMAN & \seqsplit{MAAIRKKLVIVGDGACGKTCLLIVFSKDQFPEVYVPTVFENYVADIEVDGKQVELALWDTAGQEDYDRLRPLSYPDTDVILMCFSIDSPDSLENIPEKWTPEVKHFCPNVPIILVGNKKDLRNDEHTRRELAKMKQEPVKPEEGRDMANRIGAFGYMECSAKTKDGVREVFEMATRAALQARRGKKKSGCLVL} & \textbf{PROTEIN NAME:} Epididymis secretory sperm binding protein.  \\
            \midrule
            A0A015JW94$\underline{\hspace{0.5em}}$RHIIW & \seqsplit{MANIPHGGVLKDLHARDAPKKEQLLAEVEKLPSIVLSDRQLCDLELIMNGGFSPLEGFMNQEDYQSVVNNLRLKNGLLFSMPITLDVSDQDIETLGLESKKRIVLRDPRDDAPLSILTIQDIYKPNKIEEATKVFGDDDILHPGVKYLHTQAKEFYVGGTVEAIQSPIHYDYIAHRHTPAELRAHFNKLHWTRVVAFQTRNPMHRAHRELTVRAARNRQANVLIHPVVGLTKPGDIDHYTRVRVYQALMPKYPNGMAALSLLPLAMRMGGPREAVWHAIIRKNFGNTHFIIGRDHAGPGKNSKGVDFYGPYDAQKLVAKYQDELEIEVVPFQMVTYIPDSDEYLPVDEVPEGTTTLNISGTELRRRLRNGGHIPEWFSYPEVVKVLRETHPPRSKQGFTLFLTGYYNSGKKAIGKALEVVLNQQGGRSVTLLLGETVRHGISSELGFSRRDRDQNIARIAFVSAELTKAGAAVIAAPIAPFAEARAQAKSHVETYGGFYLIHVNTPLDYCIKTDRRGIYKHNLVK} & \textbf{PROTEIN NAME:} Sulfate adenylyltransferase. \textbf{FUNCTION:} Catalyzes the first intracellular reaction of sulfate assimilation, forming adenosine-5'-phosphosulfate (APS) from inorganic sulfate and ATP. Plays an important role in sulfate activation as a component of the biosynthesis pathway of sulfur-containing amino acids. \textbf{SUBCELLULAR LOCATION:} Cytoplasm. \textbf{SIMILARITY:} Belongs to the APS kinase family. \\
            \bottomrule
    \end{tabular}\vspace{2em}
    \end{center}
\end{table*}

\section{Protein Masked Language Modeling}
To preserve unimodal information when injecting the multi-modality information from biological texts, we keep the protein Masked Language Modeling loss (MLM) widely used for PLM pre-training. Token-wise masked language modeling~\cite{BERT} reconstructs masked tokens based on the remaining context. Since the ESM-2 for initialization is pre-trained with MLM loss on protein data, we retrain MLM to avoid forgetting unimodal knowledge when injecting new information from biotext modality. Concretely, each protein residue has a 15\% chance of being masked, subsequently undergoing reconstruction based on the surrounding sequence context. $\textit{\textbf{L}}_{\text{MLM}}$ is formulated as a cross-entropy loss to measure the cost:
\begin{equation}
    \textit{\textbf{L}}_{\text{MLM}} = \mathbb{E}_{p (x^{m})} H(\Phi(x^{m}), y_x),
\end{equation}
where $H(;)$ is the cross-entropy function, $\Phi(x^{m})$ is the predicted probability of protein sequence with masked tokens $x^{m}$, and $y_x$ is the corresponding ground truth.

\section{Experiments}
\subsection{Cross-modal Transformation}
In this experiment, the raw knowledge graph data comes from PrimeKG~\cite{PrimeKG} and is in a form of triples (\textit{\{node1, node2, relation\}}). For each type of triple, we randomly sample 80\%, 10\%, and 10\% for the train, validation, and test sets, respectively. Then, we separate the test set by triple types, with a special focus on the predictions for:\textit{\{Protein, BP, Interacts with\}, \{Protein, CC, Interacts with\}, \{Protein, MF, Interacts with\}, \{Protein, Drug, Target\}, \{Disease, Protein, Associated with\}}. The metric MRR is the average reciprocal rank of all positive test triples among the corrupted negative triples. 

\subsection{Semantic Similarity Inference}
In this experiment, we obtain the groundtruth matrices from the test set released by~\cite{KeAP} where three 500 $\times$ 500 labeled matrices store the pairwise Lin Similarities of the protein associated biological process, molecular function, and cellular component. For the predicted matrices, the protein embeddings are firstly encoded from \ourapproach followed by an extra transformation module. Secondly we calculate pairwise Manhattan Similarities of the embeddings as the prediction. The higher the metric Spearman's $\rho$ between the prediction and groundtruth, the better.

\subsection{Protein-Protein Interaction Prediction}
There are totally 7 interaction types of a protein pair: \textit{reaction}, \textit{binding}, \textit{post-translational modifications (ptmod)}, \textit{activation}, \textit{inhibition}, \textit{catalysis}, and \textit{expression}. For the all three dataset, we perform Breadth-First Search (BFS) and Depth-First Search (DFS) to generate two train/validation/test splits, and only present their average results. This experiment verifies the superiority of protein representations learned by \ourapproach.

\subsection{Ablation Study on Threshold}
\begin{figure}[ht]
    \centering
    \includegraphics[width=0.92\linewidth]{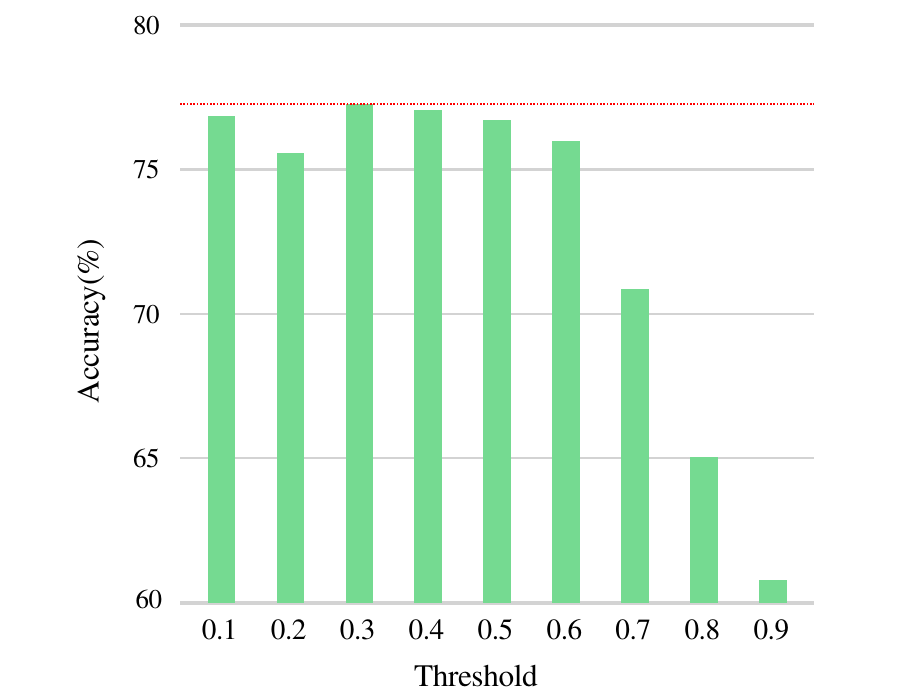}\vspace{-0.5em}
    \caption{Ablation study on different thresholds in property-grouped dynamic segment alignment loss.}
    \label{Fig:ablation_threshold}
\end{figure}

We also carry out experiments with varying values of the threshold $\theta$ in PDA. Figure~\ref{Fig:ablation_threshold} plots evaluation results on Sub dataset~\cite{DeepLoc}. We can observe that the performance exhibits fluctuations when $\theta$ is between 0.1 and 0.6, and $\theta=0.3$ is the optimal value due to the narrow advantage. When $\theta \geq 0.7$, the performance manifests a dramatic drop. The reason behind might be that the threshold is so large that too many important and functional residues are discarded. In that case, the model fails to capture any effective protein representations. Overall, we set $\theta=0.3$ in this paper.

\end{document}